\documentclass[letterpaper,10pt,conference]{ieeeconf}

\IEEEoverridecommandlockouts
\usepackage{amsmath,amssymb,amsfonts}
\usepackage{algorithm}
\usepackage{algorithmic}
\usepackage{booktabs}
\usepackage{graphicx}
\usepackage{xcolor}
\usepackage{url}
\usepackage{multirow}
\usepackage{bm}
\usepackage{cite}

\title{\LARGE \bf
CaSCo: Cascade-Aware Soft-Collision Motion Planning
}

\author{
    Shivaram Kumar$^{*}$, 
    Gaoyuan Liu,
    and Yoonchang Sung$^{\dagger}$
    \thanks{Nanyang Technological University}
    \thanks{$^{*}$This work was done while Shivaram Kumar was a visiting researcher at NTU Singapore.}
    \thanks{$^{\dagger}$Corresponding author: {\tt\small yoonchang.sung@ntu.edu.sg}}%
}

\begin{document}
\maketitle

%%%%%%%%%%%%%%%%%%%%%%%%%%%%%%%%%%%%%%%%%%%%%%%%%%%%%%%%%%%%%%%%%%%%%%%%%%%%%%%%
\begin{abstract}
Conventional motion planning treats collision as a binary constraint, although contact with different objects can have drastically different consequences.
A robot may safely brush against a cardboard box while even minor contact with a glass, laptop, or unstable object may be undesirable.
Moreover, a direct robot--object collision can move the contacted object and trigger secondary object--object collisions, making the risk of a motion depend on the physical evolution of the scene rather than only on the robot's geometric path.
We present \emph{CaSCo}, a cascade-aware soft-collision motion planning framework in which a vision-language or language model assigns semantic risk to objects and a physics simulator predicts the consequences of candidate robot motions.
CaSCo searches for a path that minimizes the total semantic risk of the unique objects displaced either directly by the robot or indirectly through cascaded collisions.
Because collisions change the environment, we augment roadmap states with the predicted object arrangement and the set of objects whose risk has already been incurred.
We develop an optimal graph-search algorithm with an admissible and consistent cascade-relaxed heuristic and caching and pruning mechanisms for efficient search.
Experiments in cluttered manipulation environments evaluate semantic risk, cascade reasoning, planning efficiency, and real-robot operation.
\vspace{-0.1cm}
\end{abstract}

%%%%%%%%%%%%%%%%%%%%%%%%%%%%%%%%%%%%%%%%%%%%%%%%%%%%%%%%%%%%%%%%%%%%%%%%%%%%%%%%
\section{Introduction}

Collision avoidance is one of the most fundamental assumptions in robot motion planning.
When an object blocks the robot's path, a conventional planner typically has two choices: find a collision-free path around it or, in manipulation settings, explicitly rearrange or remove the obstructing object before continuing.
However, neither option is always desirable.
A collision-free detour may be unnecessarily long or may not exist, while deliberately removing an obstacle requires additional manipulation planning and execution.
If incidental contact with the object is unlikely to cause meaningful harm, simply allowing the robot to push or brush against it can be substantially more efficient than planning a separate obstacle-removal action.

\begin{figure}
    \centering
    \includegraphics[width=0.75\linewidth]{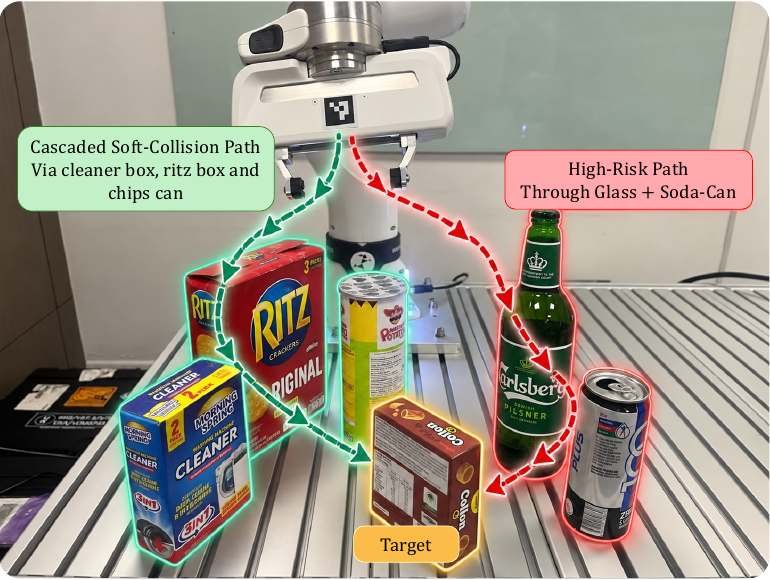}
    \vspace{-0.2cm}
    \caption{Motivating example. Two trajectories reach the target object: the left contacts more objects but incurs lower semantic risk, while the right contacts fewer objects but includes a high-risk glass bottle and a soda can that may contain liquid.}
    \label{fig:manifolds}
    \vspace{-0.7cm}
\end{figure}

These tradeoffs motivate a different view of collision in motion planning.
Rather than treating every contact as either strictly forbidden or requiring explicit object rearrangement, we consider \emph{soft collisions}: the robot may interact with movable objects when the expected consequence is acceptable.
For example, when reaching through a cluttered tabletop scene, gently displacing a lightweight cardboard box may be preferable to planning a grasp-and-place operation solely to clear it.
In contrast, contact with fragile glassware, electronics, or containers holding liquid should remain strongly discouraged.
Thus, the relevant question is not simply whether a path collides, but \emph{which objects it disturbs and how undesirable those interactions are} (Fig.~\ref{fig:manifolds}).

Recent vision-language models (VLMs) and large language models (LLMs) provide a natural mechanism for estimating such semantic differences.
For example, these models can reason that contact with a soft container is less concerning than contact with a fragile or valuable object.
Given such object-level risk values, we introduce \emph{CaSCo}, a cascade-aware soft-collision motion planning framework in which the robot seeks a path minimizing the cumulative risk of its physical interactions.
CaSCo occupies a useful middle ground between strict collision avoidance and explicit obstacle-removal planning: when contact is costly, the planner naturally avoids the object, whereas when contact is relatively harmless, it can exploit that interaction without introducing a separate manipulation subtask.

A major difficulty, however, is that collision effects can extend beyond the object directly contacted by the robot.
Pushing a low-risk object may cause it to strike a nearby fragile object.
Moreover, once an object has been displaced, subsequent robot motions occur in a different geometric environment.
Therefore, assigning a fixed cost to a roadmap edge based only on nominal robot--object intersections is insufficient.
The cost and successor state of an edge depend on the current arrangement of movable objects, which in turn depends on the preceding sequence of interactions.

CaSCo couples sampling-based motion planning with physics simulation.
We construct a probabilistic roadmap in the robot's continuous configuration space and use the simulator to evaluate the physical consequences of candidate local motions.
For each candidate motion, the simulator predicts the resulting poses of objects displaced either directly by the robot or indirectly through cascaded object--object interactions.
The search state is augmented with both the predicted object arrangement and the set of objects whose semantic risks have already been incurred.
CaSCo therefore reasons not only about whether contact occurs, but also about how that contact changes the environment and what secondary consequences it may produce.

Our formulation is closely related to Minimum Constraint Removal (MCR~\cite{hauser2014minimum}), which seeks a path that minimizes the set of violated constraints, with weighted MCR allowing different constraints to incur different costs.
In contrast, CaSCo treats movable obstacles as physical objects whose interactions can change the scene and trigger cascaded collisions.
Consequently, both future feasibility and interaction cost depend on the evolving object arrangement rather than on static obstacle labels alone.

This cascade-aware formulation produces a substantially larger search space. A search state must include not only the robot roadmap vertex, but also the current object arrangement and the set of objects whose risks have already been paid. We therefore exploit the structure of the objective to accelerate search. We introduce a cascade-relaxed heuristic that follows the simulator-predicted scene evolution while charging only directly displaced objects. Because directly displaced objects are a subset of all displaced objects, the heuristic is admissible. We further show that it is consistent. We also use lazy selective simulation to evaluate only transitions selected by promising candidate paths, together with transition caching, duplicate detection, and upper-bound pruning.

Because simulator predictions may deviate from execution, CaSCo can additionally replan when the observed object arrangement differs from its predicted arrangement beyond a predefined threshold.

The main contributions of this work are:
\begin{itemize}
    \item We introduce \emph{CaSCo}, a cascade-aware soft-collision motion planning framework that combines semantic object risk with simulated scene evolution and accounts for objects displaced through cascaded collisions.
    \item We develop an optimal graph-search method with lazy selective physics simulation, an admissible and consistent cascade-relaxed heuristic, and caching and pruning mechanisms for efficient search.
    \item Experiments demonstrate that CaSCo achieves lower cumulative semantic risk while substantially reducing planning time compared with baseline methods.
\end{itemize}

%%%%%%%%%%%%%%%%%%%%%%%%%%%%%%%%%%%%%%%%%%%%%%%%%%%%%%%%%%%%%%%%%%%%%%%%%%%%%%%%
\vspace{-0.1cm}
\section{Related Work}

CaSCo lies at the intersection of collision-tolerant motion planning, planning among movable objects, and semantic risk-aware planning. Prior work has studied how robots can relax geometric constraints, deliberately interact with movable obstacles, or use semantic knowledge to determine acceptable interactions. Our setting unifies these ideas by modeling how contact changes the scene and optimizing the semantic cost of direct and cascaded interactions.

\subsection{Collision-Tolerant and Movable-Object Planning}

MCR seeks a path that violates the smallest set of geometric constraints~\cite{hauser2014minimum,thomas2023revisiting}, while minimum constraint displacement allows obstacles to be displaced to recover feasibility~\cite{hauser2013minimum}. Weighted variants further distinguish constraints by their costs, providing a natural framework for planning when some violations are preferable to others.

Navigation and manipulation among movable obstacles instead treat environmental modification as part of planning~\cite{stilman2005navigation,nieuwenhuisen2008effective,stilman2008planning,van2009path}. Subsequent work considers physical pushing, object affordances, placement consequences, visibility, or task-level costs when determining how clutter should be modified~\cite{wang2020affordance,renault2020modeling,ellis2023navigation,zhang2023navigation,muguira2023visibility,yang2025efficient,zhang2025namo}. More generally, contact-rich planners deliberately exploit environmental interaction using optimization or sampling~\cite{posa2014direct,cheng2021contact,kang2025global}. CAT-RRT~\cite{nechyporenko2023cat} similarly permits penalized contact during sampling-based planning, while recent approaches explicitly plan motions involving movable objects~\cite{wang2025contact,ren2025collision}.

More closely related, MAMO~\cite{saxena2023planning} uses physics simulation to reason about robot--object and object--object interactions and the resulting scene evolution in clutter. CaSCo shares this dynamics-aware perspective, but objects need not be deliberately rearranged toward prescribed configurations. Instead, incidental displacement is allowed, and the planner globally minimizes a weighted, pay-once risk over objects disturbed directly or through collision cascades. Thus, CaSCo combines the global constraint-selection perspective of weighted MCR with the scene evolution modeled by MAMO.

\vspace{-0.1cm}
\subsection{Semantic Risk-Aware Planning}

Semantic knowledge has increasingly been incorporated into robot planning and safety to represent common-sense constraints and interaction preferences that geometry alone cannot capture~\cite{ikeuchi2024semantic,brunke2025semantically,kumar2026open}. Related work also uses semantic reasoning to distinguish acceptable from undesirable physical interactions. IMPACT~\cite{ling2025impact} constructs VLM-derived contact-tolerance costs for motion planning, while recent VLM-based contact-tolerant planning identifies regions where physical contact can be permitted~\cite{li2026direct}.

CaSCo extends this perspective by explicitly modeling the physical consequences of semantically weighted contact. Its cost depends not only on objects directly disturbed by the robot, but also on objects displaced indirectly through resulting object--object interactions. Because these interactions change the geometry governing future motions, the resulting object arrangement becomes part of the planning state. Consequently, identical direct contact can have different costs depending on whether it moves an object harmlessly into free space or triggers a cascade involving a high-risk object.

%%%%%%%%%%%%%%%%%%%%%%%%%%%%%%%%%%%%%%%%%%%%%%%%%%%%%%%%%%%%%%%%%%%%%%%%%%%%%%%%
\section{Problem Formulation}

\begin{figure*}[!t]
    \centering
    \includegraphics[width=1.0\textwidth]{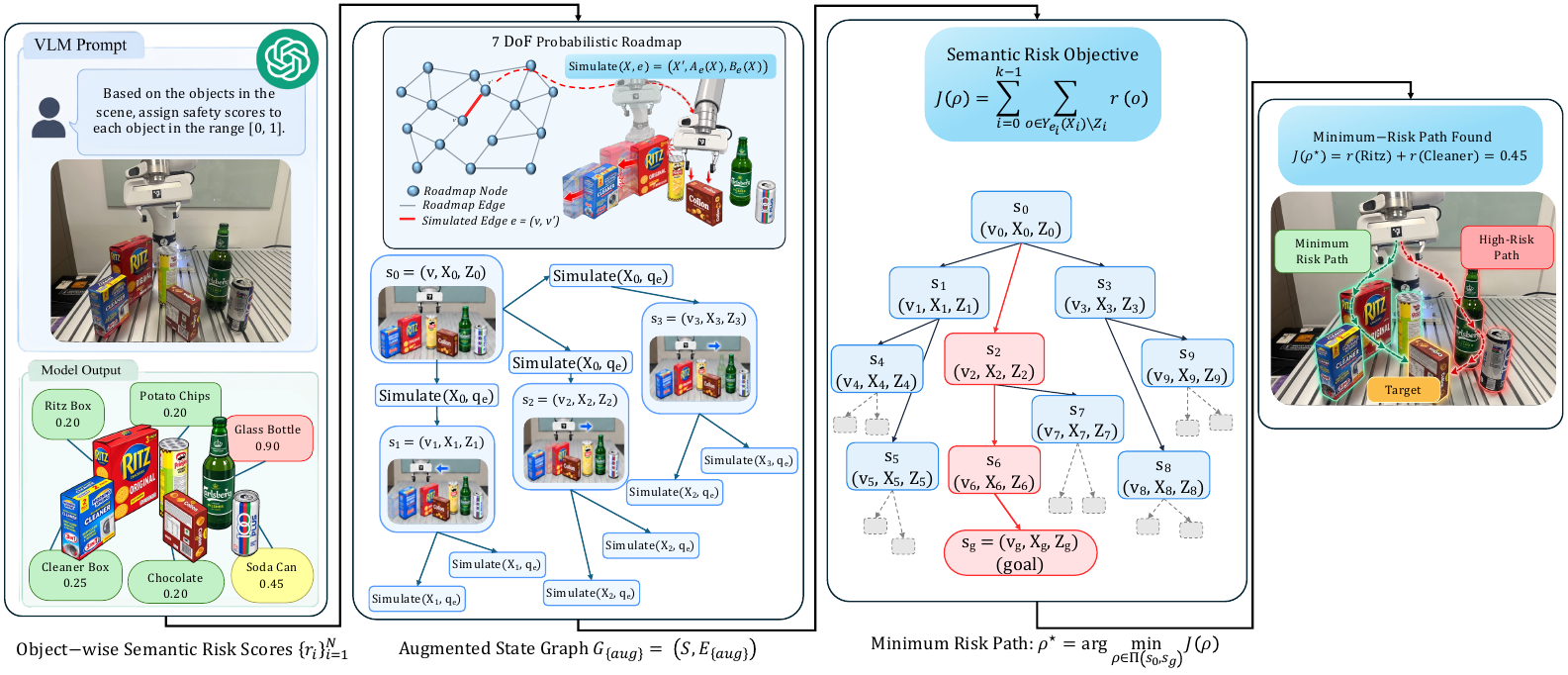}
    \vspace{-0.6cm}
    \caption{The CaSCo pipeline consists of semantic risk estimation, physics-based cascade prediction, augmented-state search, and minimum-risk planning.}
    \label{fig:architecture}
    \vspace{-0.4cm}
\end{figure*}

We formalize the soft-collision motion planning problem underlying CaSCo by combining semantic object risk, roadmap-based robot motion, and simulator-predicted cascaded interactions under a pay-once risk objective.

\subsection{Semantic Object Risk}

Let $\mathcal{O} = \{o_1,\ldots,o_m\}$ denote the set of movable objects in the scene. Each object $o_i$ is assigned a nonnegative semantic risk $r_i \geq 0$ measuring the undesirability of physically disturbing that object. For example, a lightweight cardboard container may receive low risk, while fragile glassware, electronics, or liquid-filled objects receive higher risk.

The risk values $r_i$ are provided to the planner and remain fixed during a planning episode. CaSCo is agnostic to how these values are obtained; state-dependent risk can be incorporated as an extension.

\subsection{Robot Roadmap and Scene Arrangement}

Let $G=(V,E)$ be a roadmap in robot configuration space $\mathcal{Q}$. Each vertex $v \in V$ corresponds to a robot configuration $q(v)\in\mathcal{Q}$ that is collision-free with respect to self-collision and immutable environment geometry, and an edge $e=(v,v')$ corresponds to a local trajectory $q_e:[0,1]\rightarrow\mathcal{Q}$.

Static obstacles remain hard constraints during roadmap construction, while objects in $\mathcal{O}$ are excluded from collision checks and treated as soft obstacles.

Let $X=\{x_1,\ldots,x_m\}$ denote the physical arrangement of the movable objects, where $x_i$ is the pose of object $o_i$. We assume that each simulation rollout is continued until the resulting object arrangement stabilizes before the successor state is recorded.

\subsection{Cascade-Aware Transition Model}

Executing roadmap edge $e$ from arrangement $X$ is evaluated by a physics simulator as $\operatorname{Simulate}(X,q_e)\rightarrow(X',A_e(X),B_e(X)),$ where $X'=F_e(X)$ is the resulting object arrangement. Here, $A_e(X)\subseteq\mathcal{O}$ is the set of objects directly contacted by the robot whose poses change beyond a prescribed tolerance, and $B_e(X)\subseteq\mathcal{O}\setminus A_e(X)$ is the set of objects whose poses change beyond the tolerance without direct robot contact, due to cascaded object--object interactions. The complete risk-bearing interaction set is $Y_e(X)=A_e(X)\cup B_e(X)$.

If the commanded local motion cannot be physically executed or violates a hard constraint, the simulation returns failure and the corresponding transition is discarded.

This transition is what distinguishes the proposed problem from a static weighted collision-cost formulation. Both $Y_e(X)$ and $X'$ depend on the current arrangement $X$, which itself depends on the previously executed path.

\subsection{Pay-Once Semantic Risk}

We use a pay-once objective analogous to weighted MCR. Once the semantic risk associated with an object has been incurred, subsequent interactions with the same object do not incur its risk again.

For a path $\pi=(e_1,\ldots,e_T)$ and its induced arrangement sequence $X_0,\ldots,X_T$, define its risk-bearing interaction set as
$Y(\pi)=\bigcup_{t=1}^{T}Y_{e_t}(X_{t-1})$. The semantic risk is
\begin{equation}
J(\pi)=\sum_{o_i\in Y(\pi)} r_i,
\label{eq:semantic-risk}
\end{equation}
and the goal is to find $\pi^*=\arg\min_{\pi:v_s\rightsquigarrow v_g}J(\pi)$.

Path length or execution time can optionally be included as a secondary cost, $J_\lambda(\pi)=J(\pi)+\lambda L(\pi)$, but we focus on pure semantic risk in the theoretical development.

\noindent\textbf{Proposition 1 (NP-hardness).}
Soft-collision motion planning under the pay-once risk objective is NP-hard, even in the absence of cascaded interactions.

\noindent\textit{Proof.}
We reduce from discrete MCR, which is NP-hard~\cite{hauser2014minimum}. Given a discrete MCR instance, associate each constraint with a movable object $o_i$ and assign unit risk $r_i=1$. For any roadmap edge that intersects a constraint in the MCR instance, the corresponding object is directly displaced and included in $A_e(X)$. We restrict the transition model such that these direct displacements do not trigger any secondary interactions, so $B_e(X)=\emptyset$ for every edge, and do not otherwise affect the feasibility of subsequent roadmap edges.

Under this construction, the set $Y(\pi)$ of objects displaced along a path $\pi$ is exactly the set of constraints violated by the corresponding MCR path. Therefore, $J(\pi)=\sum_{o_i\in Y(\pi)}r_i=|Y(\pi)|,$ which is precisely the discrete MCR objective. Hence, an optimal solution to this restricted soft-collision planning problem also solves the corresponding MCR instance, establishing NP-hardness. \hfill$\square$

%%%%%%%%%%%%%%%%%%%%%%%%%%%%%%%%%%%%%%%%%%%%%%%%%%%%%%%%%%%%%%%%%%%%%%%%%%%%%%%%
\section{Cascade-Aware Soft-Collision Planning}

CaSCo performs graph search over augmented states that capture the robot configuration, evolving object arrangement, and previously incurred semantic risks (Fig.~\ref{fig:architecture}).

\subsection{Augmented Search State}

A standard roadmap vertex is insufficient because two paths reaching the same robot configuration can leave different object arrangements. Moreover, because risk is paid only once, future cost also depends on which object risks have already been incurred.

We therefore define a search state as $s=(v,X,Z)$, where $v$ is the current roadmap vertex, $X$ is the predicted object arrangement, and $Z\subseteq\mathcal{O}$ is the set of objects whose risks have already been incurred.

For an edge $e=(v,v')$, simulation produces $(X',A_e,B_e)=\operatorname{Simulate}(X,q_e)$. The successor state is $s'=(v',X',Z')$, where $Z'=Z\cup A_e\cup B_e$, and the transition cost is $c(s,s')=\sum_{o_i\in(A_e\cup B_e)\setminus Z}r_i$.

A shortest path through this augmented state space therefore exactly minimizes the semantic risk objective in Eq.~\eqref{eq:semantic-risk}, subject to the roadmap discretization and simulator model.

Consider two paths that both directly touch only object $o_a$. In a static weighted-MCR formulation, they appear equivalent. Suppose, however, that the first contact pushes $o_a$ into empty space while the second pushes it into a fragile object $o_b$. Their true costs become $r_a$ and $r_a+r_b$, respectively. Furthermore, the resulting pose of $o_a$ can either clear or obstruct subsequent roadmap edges.

Thus, collision cost is neither edge-local nor determined by direct geometric intersection alone. The arrangement $X$ must therefore be part of the state and transition dynamics.

\subsection{Lazy and Selective Simulation}

Physics simulation is substantially more expensive than graph search, so CaSCo alternates between heuristic search and selective simulation rather than simulating every outgoing transition of an expanded state. During search, simulated transitions use their recorded successor arrangements and disturbance sets, with incremental cost $\Delta J=\sum_{o_i\in Y_e\setminus Z}r_i$, where $Y_e=A_e\cup B_e$. An unsimulated transition has unknown successor information, and its unevaluated suffix contributes zero additional cost as an optimistic lower bound.

The search yields a lower-bound candidate route. CaSCo simulates its first unsimulated transition using the arrangement reached by the simulated prefix. The transition record is updated with the resulting arrangement, and search is repeated. A failed simulation marks the transition infeasible.

Simulation results are cached by source arrangement and directed roadmap edge, $(X,e)$, since earlier object motion can change an edge's outcome. This restricts simulation to candidate-route transitions while reusing evaluated outcomes. A fully simulated route provides an upper bound, and search terminates when all remaining candidate lower bounds are no smaller than this cost. Thus, lazy simulation defers transition evaluation without changing the optimal solution of the fully evaluated search problem.

The main limitation is its dependence on simulator accuracy, which we mitigate during execution using discrepancy-triggered replanning.

\subsection{Cascade-Relaxed MCR Heuristic}

The augmented state space can be substantially larger than the original roadmap, as states differ in both object arrangement $X$ and paid set $Z$. To improve search efficiency, we introduce a cascade-relaxed MCR heuristic that exploits the structure of the pay-once objective.

Let $h^*(s)$ denote the optimal remaining semantic risk from state $s=(v,X,Z)$. We define a cascade-relaxed heuristic that preserves the simulator-predicted evolution of the object arrangement while accounting only for objects displaced directly by the robot and ignoring the risk of objects displaced through cascaded interactions. For any roadmap path $\rho:v\rightsquigarrow v_g$, let $X_e^\rho$ denote the simulator-predicted object arrangement immediately before executing edge $e$ along $\rho$. The heuristic is then 
\begin{equation} 
h_{\mathrm{MCR}}(s) = \min_{\rho:v\rightsquigarrow v_g} \sum_{o_i\in \left(\bigcup_{e\in\rho}A_e(X_e^\rho)\right)\setminus Z} r_i. 
\label{eq:mcr-heuristic} 
\end{equation}

Importantly, the heuristic does not freeze the scene: cascaded interactions may change the arrangement and thereby affect future direct contacts. The relaxation only ignores the semantic costs associated with the cascade sets $B_e$. Under lazy simulation, cached transitions use their simulated successor arrangements and direct-displacement sets, while the cost beyond the first unsimulated transition is set to zero. Since omitted risks are nonnegative, this yields a valid lower bound and is updated as the transition cache expands.

\textit{Admissibility:} A heuristic is admissible if it never overestimates the optimal remaining cost, i.e., $h_{\mathrm{MCR}}(s)\leq h^(s)$. For every edge and arrangement, $A_e(X)\subseteq A_e(X)\cup B_e(X)=Y_e(X)$. Since $r_i\geq0$, considering only directly displaced objects cannot exceed the true cascade-aware cost. Lazy evaluation can only further decrease this estimate by omitting unsimulated suffix costs. Therefore, $h_{\mathrm{MCR}}(s)\leq h^(s)$, and the heuristic remains admissible.

\textit{Consistency:} A heuristic is consistent if, for every transition $s\rightarrow s'$, it satisfies $h_{\mathrm{MCR}}(s)\leq c(s,s')+h_{\mathrm{MCR}}(s')$. Consider a transition from $s=(v,X,Z)$ to $s'=(v',X',Z\cup Y_e(X))$. Prepending $e$ to a path achieving $h_{\mathrm{MCR}}(s')$ yields a valid path from $s$. Any directly displaced object contributing to this path is either newly incurred on edge $e$ and covered by $c(s,s')=\sum_{o_i\in Y_e(X)\setminus Z}r_i$, or remains unpaid and is accounted for in $h_{\mathrm{MCR}}(s')$. Hence, $h_{\mathrm{MCR}}(s)\leq c(s,s')+h_{\mathrm{MCR}}(s')$, establishing consistency.

\subsection{Pareto-Frontier Heuristic Cache}

Exact evaluation of Eq.~\eqref{eq:mcr-heuristic} itself requires solving a cascade-relaxed, pay-once search problem and can therefore be expensive. We reduce this overhead by exploiting the fact that candidate direct-contact sets depend on $(v,X)$ but not on the paid set $Z$. Consequently, heuristic information computed for a given $(v,X)$ can be reused across search states that differ only in $Z$.

For each roadmap path $\rho:v\rightsquigarrow v_g$, let $U_\rho=\bigcup_{e\in\rho}A_e(X_e^\rho)$ denote its direct-displacement set. Then $h_{\mathrm{MCR}}(v,X,Z)=\min_\rho\sum_{o_i\in U_\rho\setminus Z}r_i$. If $U_{\rho_1}\subset U_{\rho_2}$, then $\rho_2$ can never have lower discounted risk for any $Z$. We therefore retain only subset-minimal direct-displacement sets, forming a Pareto frontier $\mathcal{P}(v,X)$, so that $h_{\mathrm{MCR}}(v,X,Z)=\min_{U\in\mathcal{P}(v,X)}\sum_{o_i\in U\setminus Z}r_i$.

The frontier can still be exponential in the worst case. When exact heuristic evaluation is prohibitive, setting $h=0$ recovers Dijkstra search without affecting correctness. We therefore treat the Pareto-frontier heuristic as an acceleration mechanism rather than a requirement for optimality.

\subsection{Upper-Bound Pruning and Duplicate Detection} 

Any feasible solution provides an upper bound $\mathrm{UB}$. By admissibility, any state $s$ satisfying $g(s)+h_{\mathrm{MCR}}(s)\geq\mathrm{UB}$ can be pruned while retaining the current best solution.

For duplicate detection, we compare the full augmented state $(v,X,Z)$, including the robot vertex, object arrangement, and previously incurred risks. If the same state is reached through multiple paths, only the instance with the smallest cost-to-come $g(s)$ is retained, while higher-cost duplicates are discarded.

\subsection{Search Algorithm}

Algorithm~\ref{alg:cascade-search} summarizes the alternating search-and-simulation procedure. Each search uses cached transition outcomes and assigns zero lower-bound cost to unsimulated transitions without assuming their successor arrangements. The first unsimulated transition on the minimum-lower-bound candidate route is then evaluated, and search is repeated. A fully simulated route is retained as the current best solution, and the algorithm terminates when no remaining candidate can improve its cost.

\begin{algorithm}[t]
\caption{CaSCo Lazy Graph Search}
\label{alg:cascade-search}
\begin{algorithmic}[1]
\STATE $s_0\gets(v_s,X_0,\emptyset)$, $\mathcal{C}\gets\emptyset$,
       $\mathrm{UB}\gets\infty$, $\pi^*\gets\mathrm{failure}$
\WHILE{\TRUE}
\STATE $\pi\gets\operatorname{Search}(s_0,\mathcal{C},h_{\mathrm{MCR}},\mathrm{UB})$
\IF{$\pi=\emptyset$}
    \RETURN $\pi^*$
\ENDIF
\STATE $(X,e)\gets\operatorname{FirstUnsimulated}(\pi,\mathcal{C})$
\IF{no such transition exists}
    \STATE $\mathrm{UB}\gets J(\pi)$, $\pi^*\gets\pi$
    \STATE \textbf{continue}
\ENDIF
\STATE $\mathcal{C}[X,e]\gets\operatorname{Simulate}(X,q_e)$
\ENDWHILE
\end{algorithmic}
\end{algorithm}

Within $\operatorname{Search}$, cached successful transitions use their recorded successor arrangements and disturbance sets, while cached failures are excluded. An unsimulated transition does not instantiate a successor state; its remaining suffix instead contributes zero additional cost as a lower bound. Search returns the minimum-lower-bound candidate below $\mathrm{UB}$, or no candidate if none can improve the current best solution.

\subsection{Discrepancy-Triggered Replanning}

Because simulator-predicted object motions may deviate from execution, we use a simple replanning safeguard. After executing each roadmap edge, CaSCo compares the observed object arrangement $\hat{X}_{k+1}$ with its predicted arrangement $X_{k+1}^{\mathrm{pred}}$. If $d(\hat{X}_{k+1},X_{k+1}^{\mathrm{pred}})>\delta,$ for a predefined scene-discrepancy threshold $\delta$, CaSCo replans from the observed state; otherwise, execution continues along the current plan.

%%%%%%%%%%%%%%%%%%%%%%%%%%%%%%%%%%%%%%%%%%%%%%%%%%%%%%%%%%%%%%%%%%%%%%%%%%%%%%%%
\section{Semantic Risk Estimation}
\label{sec:risk}

CaSCo requires a semantic risk weight $r_i$ for each movable object $o_i$ but is agnostic to how these weights are obtained.
In our experiments, we instantiate this module using a VLM based semantic risk estimator.

Rather than directly requesting arbitrary numerical values, the model assigns each object $o_i$ an ordinal risk level $\ell_i\in\{0,\ldots,4\}$ corresponding to $\{\text{negligible},\text{low},\text{medium},\text{high},\text{critical}\}$. We map these levels to fixed nonnegative weights $\{w_0,\ldots,w_4\}$ and set $r_i=w_{\ell_i}$. This provides a more interpretable interface and reduces sensitivity to arbitrary numerical outputs from the model. These levels remain soft costs; objects that must never be disturbed are instead represented as hard constraints.

The prompt evaluates the consequence of physically disturbing an object rather than its value alone. For example, an empty plastic cup and a cup containing hot liquid may belong to the same object category but receive different risk levels because disturbing them can have different consequences.

For reproducibility, semantic risks are estimated before planning and remain fixed throughout each planning episode. The risk-estimation module can therefore be replaced independently of the planning algorithm.

%%%%%%%%%%%%%%%%%%%%%%%%%%%%%%%%%%%%%%%%%%%%%%%%%%%%%%%%%%%%%%%%%%%%%%%%%%%%%%%%
\section{Experiments}

We evaluate CaSCo through simulation and real-robot experiments designed to answer three questions:
\begin{enumerate}
    \item \textbf{Semantic and dynamic planning:} Does CaSCo reduce realized semantic risk relative to geometric, uniform-cost, and static semantic planning when contact is permitted?
    \item \textbf{Cascade reasoning:} Does accounting for secondary object--object interactions reduce realized semantic risk and cascade disturbances compared with dynamic planning that considers only direct robot contact?
    \item \textbf{Search efficiency:} Does the proposed cascade-relaxed heuristic reduce planning time, state exploration, and physics simulation while preserving solution quality relative to Dijkstra search?
\end{enumerate}

\subsection{Experimental Setup}

We evaluate CaSCo in MuJoCo using a Franka manipulator in two cluttered environment families: shelf reaching and tabletop manipulation. Each family contains 20 independently sampled instances with varying roadmaps and clutter arrangements. Each roadmap contains 1,200 sampled nodes and 15 diversified routes. For each instance, all methods share the same roadmap, initial object arrangement, and semantic risks, so differences reflect the planning formulation rather than geometric or semantic variation. The primary evaluation therefore comprises 280 instance--method runs with a $300,\mathrm{s}$ search budget.

\begin{figure}[t]
    \centering
    \includegraphics[width=0.91\columnwidth]{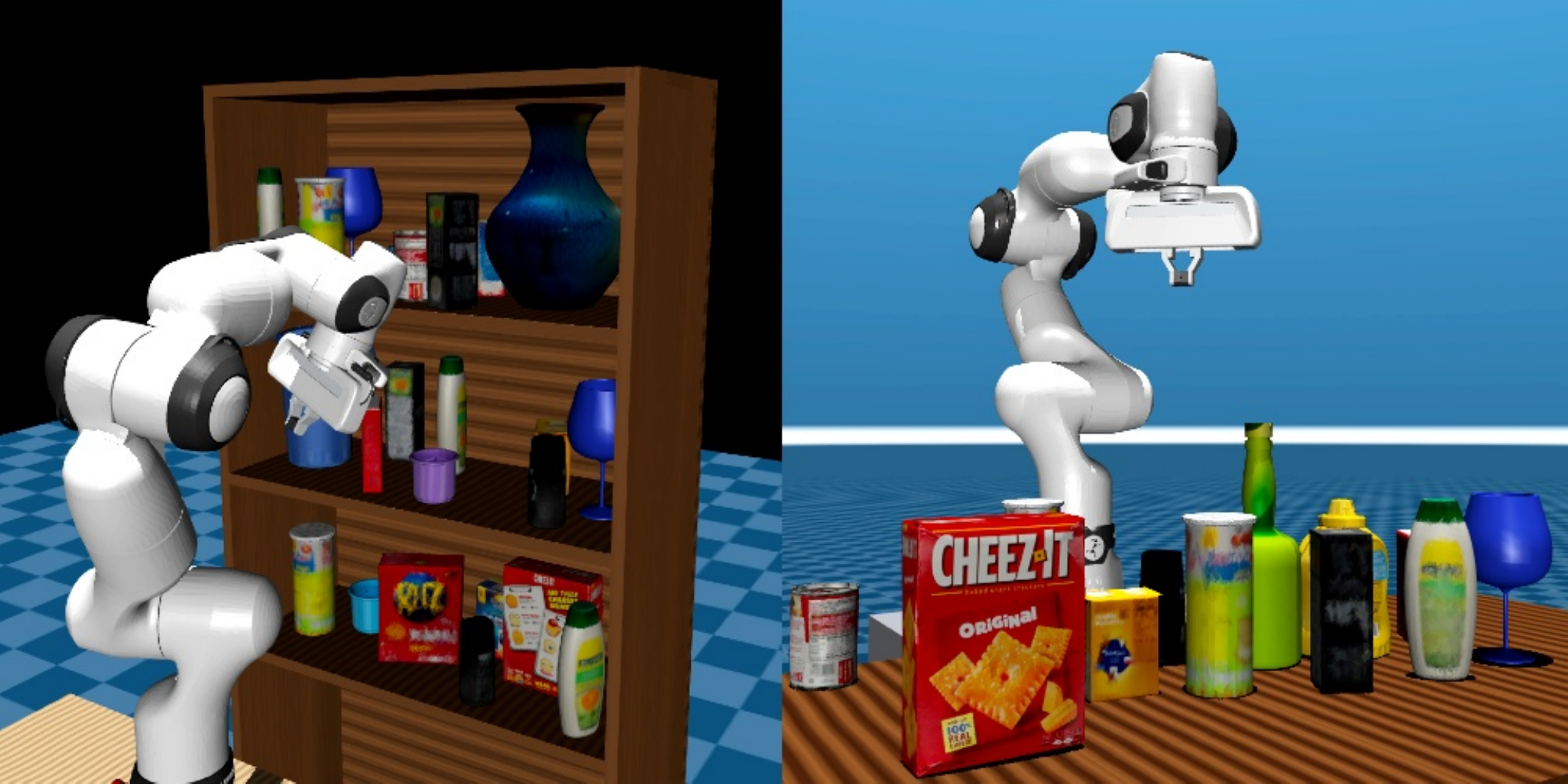}
    \vspace{-0.2cm}
    \caption{Representative MuJoCo environments used for evaluation:
    (a) shelf reaching and (b) tabletop manipulation. Roadmaps and
    clutter arrangements are independently resampled across instances.}
    \label{fig:evaluation_environments}
    \vspace{-0.3cm}
\end{figure}

For every method, the returned path is independently replayed from the initial scene in a fresh physics simulation, and a run is successful only if the robot reaches the goal. Replay provides realized semantic risk, direct and cascade-only disturbances, and total object displacement. Search efficiency is measured using planning time, distinct augmented states, and physics rollouts performed during planning; evaluation replays are not included in rollout counts.

These metrics capture complementary aspects of planner behavior. Semantic risk is the primary objective, while direct and cascade-only disturbances reveal how that risk is incurred. We additionally report total object displacement as a secondary measure of scene disturbance. A planner may intentionally make more direct contacts while achieving lower overall risk by avoiding costly secondary interactions.

\subsection{Baselines}

We compare five baselines and two CaSCo search variants: \textbf{Collision-Free} treats movable objects as hard obstacles; \textbf{Shortest Soft} permits contact but minimizes path length without semantic risk; \textbf{Uniform-MCR} uses unit object costs with frozen geometric edge labels; \textbf{Weighted Static MCR} uses CaSCo's semantic weights but ignores scene evolution; \textbf{Direct-Only Dynamic} updates the scene via physics simulation but charges only directly disturbed objects; \textbf{CaSCo-Dijkstra} optimizes the full dynamic, cascade-aware pay-once objective with $h=0$; and \textbf{CaSCo-$h_{\mathrm{MCR}}$} optimizes the same objective using $\mathrm{A^*}$ with the proposed heuristic. 

These baselines progressively isolate the components of the proposed formulation. Uniform-MCR removes semantic weighting, Weighted Static MCR introduces semantic weighting while keeping the environment fixed, and Direct-Only Dynamic further models scene evolution while excluding the semantic cost of cascade-only disturbances. Thus, the static baselines test whether assigning more informative object costs is sufficient without modeling physical consequences, whereas Direct-Only Dynamic tests whether scene evolution alone is sufficient without reasoning about secondary interactions. The comparison between Direct-Only Dynamic and CaSCo therefore isolates the effect of cascade-aware reasoning. The two CaSCo variants share the same supporting search mechanisms and differ only in heuristic guidance, allowing us to isolate the proposed heuristic as the primary search contribution.

The baselines also distinguish three increasingly expressive contact models: fixed geometric labels, dynamic direct-contact consequences, and dynamic direct-plus-cascade consequences. A static planner may assign the correct semantic cost to the initially contacted object yet still choose a poor path if that interaction moves it into another high-risk object. The dynamic baselines therefore test whether modeling scene evolution, rather than only improving object-level costs, is necessary for reliable soft-collision planning.

\subsection{Semantic and Cascade-Aware Planning}

Table~\ref{tab:main_quality} summarizes the realized outcomes. Collision-Free produces no usable plan in either environment, indicating that interaction with movable clutter is necessary for the tasks. In contrast, both CaSCo variants solve all evaluated instances.

\begin{table}[t]
\centering
\caption{Outcomes over 20 instances per environment. Quality metrics are medians over successful replays.}
\vspace{-0.2cm}
\label{tab:main_quality}
\footnotesize
\setlength{\tabcolsep}{2pt}
\resizebox{\columnwidth}{!}{
\begin{tabular}{@{}lccccc@{}}
\toprule
\textbf{Method} &
\textbf{Success (\%)} &
\textbf{Risk} $\downarrow$ &
\textbf{Direct} &
\textbf{Cascade} $\downarrow$ &
\textbf{Disp. (m)} $\downarrow$ \\
\midrule
\multicolumn{6}{c}{\textbf{Shelf}} \\
Collision-Free            & 0   & --   & -- & -- & --    \\
Shortest Soft             & 50  & 28.0 & 3  & 5  & 0.426 \\
Uniform-MCR               & 85  & 26.0 & 2  & 6  & 0.424 \\
Weighted Static MCR       & 85  & 24.0 & 2  & 6  & 0.424 \\
Direct-Only Dynamic       & 95  & 26.0 & 2  & 6  & 0.386 \\
CaSCo-Dijkstra            & 100 & 18.5 & 3  & 3  & 0.338 \\
CaSCo-$h_{\mathrm{MCR}}$  & 100 & 18.5 & 3  & 3  & 0.338 \\
\midrule
\multicolumn{6}{c}{\textbf{Tabletop}} \\
Collision-Free            & 0   & --   & -- & -- & --    \\
Shortest Soft             & 50  & 27.0 & 2  & 4  & 0.254 \\
Uniform-MCR               & 55  & 25.0 & 2  & 4  & 0.310 \\
Weighted Static MCR       & 70  & 42.0 & 2  & 7  & 0.482 \\
Direct-Only Dynamic       & 95  & 38.0 & 2  & 7  & 0.498 \\
CaSCo-Dijkstra            & 100 & 25.0 & 2  & 3  & 0.260 \\
CaSCo-$h_{\mathrm{MCR}}$  & 100 & 25.0 & 2  & 3  & 0.260 \\
\bottomrule
\vspace{-0.4cm}
\end{tabular}}
\end{table}

On shelf, CaSCo reduces median realized risk to $18.5$, compared with $28.0$ for Shortest Soft, $26.0$ for Uniform-MCR and Direct-Only Dynamic, and $24.0$ for Weighted Static MCR. Direct-Only Dynamic also produces six cascade-only disturbances, compared with three for CaSCo. Although CaSCo may directly disturb more objects than some baselines, this is consistent with its objective: the planner minimizes not contact count, but the semantic consequence of the interaction sequence. Direct contact with a low-risk object can therefore be preferable when it avoids more costly downstream interactions.

The tabletop family more clearly exposes the limitation of static or direct-only reasoning. Weighted Static MCR incurs median risk $42.0$ despite using semantic object weights, while Direct-Only Dynamic obtains risk $38.0$, seven cascade-only disturbances, and $0.498\,\mathrm{m}$ displacement. CaSCo instead achieves risk $25.0$, three cascade-only disturbances, and $0.260\,\mathrm{m}$ displacement while solving all instances. The poor performance of Weighted Static MCR is particularly informative: a path that appears preferable under the initial arrangement can become undesirable once contacted objects move and alter subsequent interactions. These results indicate that semantic weighting alone is insufficient when contact changes the scene, and that accounting for the secondary physical consequences of contact can substantially alter which trajectory is preferable.

Because randomly sampled instances can differ substantially in difficulty, we perform paired comparisons on matched instances using CaSCo-$h_{\mathrm{MCR}}$ as the reference. For each metric and instance $i$, we compute $d_i=\mathrm{Baseline}_i-\mathrm{CaSCo}_i$ and report the mean paired difference with a 95\% confidence interval in Table~\ref{tab:quality_ci}. For lower-is-better metrics, a confidence interval entirely above zero indicates a statistically significant improvement by CaSCo. Pairing compares methods on the same sampled planning problem and therefore reduces the influence of differences in intrinsic instance difficulty. Direct-contact counts are reported descriptively and are not interpreted as a lower-is-better objective.

\begin{table*}[t]
\centering
\caption{Paired mean differences relative to CaSCo-$h_{\mathrm{MCR}}$ with 95\% confidence intervals, where $d_i=\mathrm{Baseline}_i-\mathrm{CaSCo}_i$. For Risk, Cascade, and Disp., intervals entirely above zero favor CaSCo.}
\vspace{-0.2cm}
\label{tab:quality_ci}
\footnotesize
\setlength{\tabcolsep}{4pt}
\begin{tabular}{@{}llcccc@{}}
\toprule
\textbf{Scene} & \textbf{Baseline} &
\textbf{Risk} &
\textbf{Direct} &
\textbf{Cascade} &
\textbf{Disp. (m)} \\
\midrule
\multirow{4}{*}{Shelf}
& Shortest Soft
& $+9.20\;(4.4,14.2)$
& $+1.00\;(0.3,1.8)$
& $+2.00\;(0.7,3.5)$
& $+0.166\;(0.06,0.28)$ \\
& Uniform-MCR
& $+7.80\;(5.3,10.2)$
& $-0.53\;(-0.93,-0.13)$
& $+2.87\;(1.9,3.9)$
& $+0.123\;(0.05,0.20)$ \\
& Weighted Static MCR
& $+7.31\;(4.8,9.8)$
& $-0.38\;(-0.81,0.13)$
& $+2.44\;(1.5,3.4)$
& $+0.099\;(0.03,0.17)$ \\
& Direct-Only Dynamic
& $+7.16\;(4.5,10.0)$
& $-0.42\;(-0.79,0.00)$
& $+2.42\;(1.5,3.4)$
& $+0.098\;(0.03,0.16)$ \\
\midrule
\multirow{4}{*}{Tabletop}
& Shortest Soft
& $+7.00\;(4.1,16.8)$
& $-0.30\;(-0.7,0)$
& $+2.10\;(0,4.8)$
& $+0.169\;(0,0.39)$ \\
& Uniform-MCR
& $+5.64\;(5.1,13.7)$
& $-0.27\;(-0.64,0)$
& $+1.55\;(0,3.6)$
& $+0.116\;(0,0.27)$ \\
& Weighted Static MCR
& $+12.20\;(3.3,22.5)$
& $+0.30\;(0,0.7)$
& $+2.80\;(0.8,5.2)$
& $+0.184\;(0.03,0.37)$ \\
& Direct-Only Dynamic
& $+11.82\;(3.1,22.0)$
& $-0.35\;(-0.71,0)$
& $+3.47\;(0.94,6.35)$
& $+0.284\;(0.05,0.56)$ \\
\bottomrule
\vspace{-0.3cm}
\end{tabular}
\end{table*}

The paired results reinforce the aggregate trends. Semantic-risk intervals are entirely above zero for every baseline. In particular, Direct-Only Dynamic has higher realized risk than CaSCo by $7.16$ on shelf and $11.82$ on tabletop on average, while also producing significantly more cascade disturbances. Since both methods propagate the scene through physics simulation, this comparison provides the clearest evidence that explicitly incorporating cascade consequences into the planning objective improves the selected interactions rather than merely benefiting from dynamic scene updates. The displacement results show the same trend, indicating that avoiding high-risk cascades also tends to reduce the overall physical disturbance of the scene.

\vspace{-0.1cm}
\subsection{Search Acceleration}
\vspace{-0.1cm}

CaSCo-Dijkstra and CaSCo-$h_{\mathrm{MCR}}$ optimize the same objective using the same transition model, allowing us to isolate the computational effect of heuristic-guided search. Table~\ref{tab:search_efficiency} compares their planning time, explored states, and physics rollouts. We again use paired differences, now defined as $d_i=\mathrm{Dijkstra}_i-\mathrm{A^*}_i$, so positive differences indicate lower computational cost for $\mathrm{A^*}$.

\begin{table}[t]
\centering
\caption{Search efficiency of Dijkstra and $\mathrm{A^*}$ with $h_{\mathrm{MCR}}$. $\Delta$ denotes the paired difference Dijkstra $-$ $\mathrm{A^*}$.}
\vspace{-0.2cm}
\label{tab:search_efficiency}
\footnotesize
\setlength{\tabcolsep}{2.5pt}
\resizebox{\columnwidth}{!}{
\begin{tabular}{@{}llrrc@{}}
\toprule
\textbf{Scene} & \textbf{Metric} &
\textbf{Dijkstra} & \textbf{$\mathrm{A^*}$} &
\textbf{$\Delta$ mean (95\% CI)} \\
\midrule
Shelf
& Time (s)    & 31.28 & 10.69
& $+22.06\;(17.42,26.70)$ \\
& States      & 712 & 219
& $+154.20\;(221.8, 524.9)$ \\
& Rollouts    & 122 & 31.5
& $+83.70\;(17.50,170.90)$ \\
\midrule
Tabletop
& Time (s)    & 121.96 & 26.13
& $+12.46\;(7.61,18.91)$ \\
& States      & 4,514.5 & 665
& $+651.10\;(327.2, 1038.7)$ \\
& Rollouts    & 1,348 & 103
& $+398.82\;(185.35,661.41)$ \\
\bottomrule
\vspace{-0.3cm}
\end{tabular}}
\end{table}

On shelf, $\mathrm{A^*}$ reduces median planning time from $31.28\,\mathrm{s}$ to $10.69\,\mathrm{s}$, distinct states from 712 to 219, and physics rollouts from 122 to 31.5, corresponding to reductions of approximately $66\%$, $69\%$, and $74\%$. On tabletop, these quantities decrease from $121.96\,\mathrm{s}$ to $26.13\,\mathrm{s}$, 4,514.5 to 665 states, and 1,348 to 103 rollouts, corresponding to reductions of approximately $79\%$, $85\%$, and $92\%$. The paired confidence intervals for planning time and physics rollouts remain above zero in both families, showing that the reduction persists when the methods are compared on matched instances.

The reduction in physics rollouts is important because each rollout requires evaluating the physical consequence of a candidate local motion. Thus, the heuristic improves performance not only by visiting fewer augmented states, but also by avoiding expensive simulator queries associated with less promising portions of the search space. Both search variants solve all tabletop instances and recover the same route and replayed semantic risk across the trials, indicating that the computational savings do not come from changing the planning objective or accepting lower-quality solutions.

We additionally sample 20 roadmap realizations per environment over fixed object layouts to test whether the observed gain depends on a particular roadmap. Both variants again solve every sampled realization, while $\mathrm{A^*}$ consistently requires less computation. On the shelf sweep, median planning time decreases from $13\,\mathrm{s}$ to $8\,\mathrm{s}$ and physics rollouts from 36 to 23, while both methods obtain median replayed risk $17$. The reduction is smaller than in the primary benchmark, as expected because only the roadmap varies while the clutter is fixed, but the qualitative advantage remains consistent. These additional trials provide evidence that the benefit of $h_{\mathrm{MCR}}$ persists across different sampled motion graphs rather than arising from a particular roadmap realization.

\subsection{Real-Robot Demonstrations}

We finally deploy CaSCo on a physical Franka arm to demonstrate end-to-end operation beyond the simulated benchmarks, as shown in Fig.~\ref{fig:manifolds}. Starting from observations of the workspace, we construct a planning scene by recovering object geometry and pose and associate each movable object with a semantic interaction cost. The reconstructed scene is then used by the same physics-based transition model employed by the planner, allowing candidate robot motions to be evaluated together with their predicted effects on surrounding objects before execution. CaSCo executes the resulting minimum-risk motion on the physical robot.

We evaluate cluttered scenes with both benign contacts and cases where moving one object induces secondary motion of nearby objects. The supplementary video shows representative executions in which the robot may contact a low-risk object to avoid cascades involving more sensitive objects. These demonstrations qualitatively validate end-to-end CaSCo operation on real hardware using perception-derived scene models. The implementation and real-robot code will be released upon publication.

\section{Conclusion}

We present \emph{CaSCo}, a cascade-aware soft-collision motion planning framework that replaces binary collision avoidance with semantic risk-aware interaction. By combining semantic object risks with physics-based prediction, CaSCo accounts for both direct contacts and cascaded effects on the evolving scene. We formulate planning as shortest-path search over augmented states and develop an admissible and consistent cascade-relaxed heuristic with efficient pruning. Experiments in simulation and on a real robot show that CaSCo reduces realized semantic risk while maintaining practical planning efficiency. Future work will address uncertainty and scalability to more complex interactions.

% \section{Acknowledgment}
% The authors acknowledge the use of Gemini to visually enhance some illustrations in Figs.~\ref{fig:manifolds} and~\ref{fig:architecture} without altering their scientific content.

%%%%%%%%%%%%%%%%%%%%%%%%%%%%%%%%%%%%%%%%%%%%%%%%%%%%%%%%%%%%%%%%%%%%%%%%%%%%%%%%
% NOTE:
% Replace/verify bibliographic metadata with official BibTeX before submission.
% These entries are included to keep this draft self-contained.

\bibliographystyle{IEEEtran}
\bibliography{references}

\end{document}